\documentclass{article}
\usepackage{graphicx}
\usepackage{url}
\usepackage{amsmath}
\usepackage[utf8]{inputenc}
\usepackage{multirow}
\usepackage{hyperref}
\usepackage{authblk}
\usepackage[numbers]{natbib}
\usepackage{xcolor}
\usepackage{booktabs}
\usepackage{booktabs}
\usepackage{multirow}
\usepackage{threeparttable}
\usepackage{makecell} % for \makecell or \shortstack
\usepackage{booktabs}
\usepackage{multirow}
\usepackage{threeparttable}
\usepackage{makecell}
\usepackage{graphicx}

\title{BreastDCEDL: A Comprehensive Breast Cancer DCE-MRI Dataset and Transformer Implementation for Treatment Response Prediction}

\author{
  Naomi Fridman\textsuperscript{1,*}, 
  Bubby Solway\textsuperscript{2}, 
  Tomer Fridman\textsuperscript{2}, 
  Itamar Barnea\textsuperscript{2}, 
  Anat Goldstein\textsuperscript{1} \\
  \textsuperscript{1}Dept. of Industrial Engineering and, Ariel University, Ariel, Israel \\
  \textsuperscript{2}NF Algorithms \& AI, Tel Aviv, Israel \\
  \textsuperscript{*}Corresponding author: \texttt{naominoe.fridman@msmail.ariel.ac.il}
}

\date{}

\begin{document}
\maketitle

\begin{abstract}

Breast cancer is a leading cause of cancer-related mortality worldwide, making early detection and accurate treatment response monitoring critical. We present BreastDCEDL, a curated, deep learning-ready dataset comprising pretreatment 3D Dynamic Contrast-Enhanced MRI (DCE-MRI) scans from 2,070 breast cancer patients drawn from the I-SPY1, I-SPY2 and Duke cohorts, all sourced from The Cancer Imaging Archive. The raw DICOM imaging data was converted into standardized 3D NIfTI volumes, preserving the integrity of the signal. Unified tumor annotations and harmonized clinical metadata, including pathologic complete response (pCR), hormone receptor (HR), and HER2 status, are also provided. DCE-MRI provides essential diagnostic information and deep learning offers potential for analyzing such complex data; however, progress has been limited by a lack of accessible, public, multicenter datasets. BreastDCEDL addresses this gap, enabling the development of advanced models, including state-of-the-art transformer architectures that require substantial training data. To demonstrate its capacity for robust modeling, we developed the first transformer-based model for breast DCE-MRI, utilizing a Vision Transformer (ViT) architecture trained on RGB-fused images from three contrast phases: precontrast, early postcontrast, and late postcontrast. Our ViT model achieved state-of-the-art pCR prediction performance in HR+/HER2$^-$ patients, with an AUC of 0.94 and an accuracy of 0.93. BreastDCEDL includes predefined benchmark splits, offering a framework for reproducible research and enabling clinically meaningful modeling in breast cancer imaging.

\end{abstract}

\section{Introduction}

Breast cancer is a leading cause of cancer death worldwide; Early detection and accurate treatment monitoring are crucial to better patient outcomes. An important clinical decision after diagnosis is the choice between immediate surgery or neoadjuvant chemotherapy (NAC) before surgery. NAC aims to reduce tumor size, enhancing the potential for surgical success. The optimal outcome is the complete pathologic response (pCR), defined as the absence of residual invasive disease following therapy, confirmed by postsurgical biopsy. Accurate prediction of pCR is critical because it strongly correlates with improved long-term survival and can guide personalized treatment decisions.

\begin{figure}[h!]
\centering
\includegraphics[width=\textwidth]{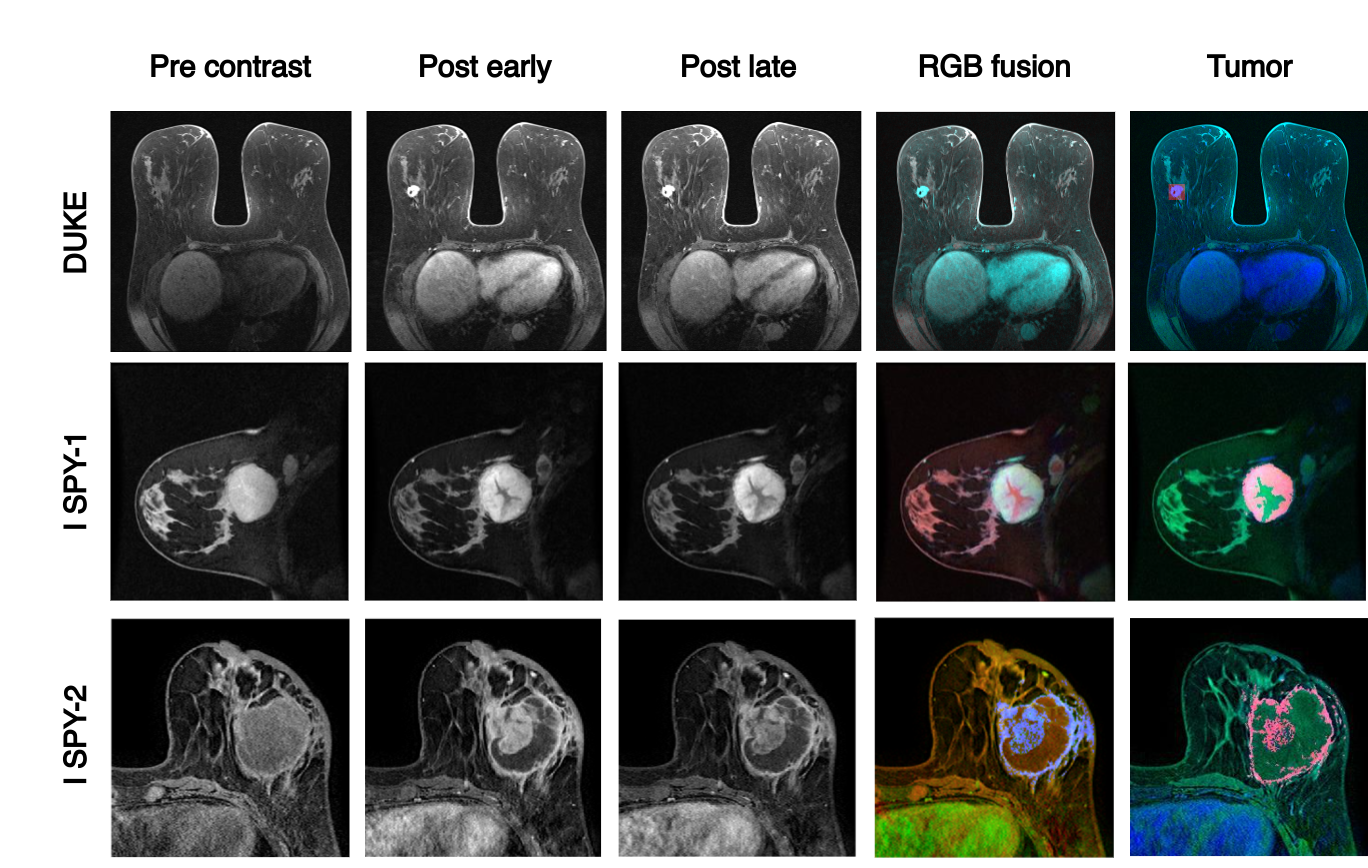}
\caption{Dynamic Contrast-Enhanced MRI illustrates tumor segmentation across datasets. Representative DCE-MRI from the Duke, I-SPY1, and I-SPY2 datasets show pre-contrast, early post-contrast, and late post-contrast acquisitions. Columns display individual phases (left), a merged RGB composite (center), and tumor segmentation (right). Tumor localization in the Duke dataset is highlighted by a bounding box.}
\label{fig:dce_mri_visualization}
\end{figure}

Dynamic contrast-enriched magnetic resonance imaging (DCE-MRI) is a principal imaging modality for the evaluation of breast cancer. The technique involves acquiring sequential 3D scans before and after the administration of contrast agents, typically analyzed at three time points: baseline before contrast, early post-contrast (maximum enhancement) and late post-contrast (washout dynamics), as shown in Figure~\ref{fig:dce_mri_visualization}. These acquisitions enable visualization of abnormal tumor vascularity, characterized by increased vessel density, permeability, and disorganized structure, which distinguishes malignant tissues from normal breast parenchyma.

Deep learning applied to breast DCE-MRI analysis has shown promise, but faces limitations. Multiple systematic reviews~\cite{Gullo_Eskreis-Winkler_Morris_Pinker_2019, Houssein_Emam_Ali_Suganthan_2021, Wei_Ma_Liu_2024} have identified critical barriers: insufficient access to large-scale datasets, inconsistent evaluation protocols, and inadequate external validation. Although convolutional neural networks have shown promising results through multiphase fusion~\cite{Liu_Mutasa_Chang_Siddique_Jambawalikar_Ha_2020}, multimodal integration~\cite{Dammu_Ren_Duong_2023, Joo_Ko_Kwon_Jeon_Jung_Kim_Chung_Im_2021}, and multi-institutional validation~\cite{Braman_2020}, most investigations remain constrained by limited sample sizes and non-standardized data formats.

Transformer architectures, which have revolutionized natural language processing and computer vision~\cite{Vaswani_Shazeer_Parmar_Uszkoreit_Jones_Gomez_Kaiser_Polosukhin_2017}, offer substantial potential for medical imaging applications~\cite{Li_Chen_Tang_Wang_Landman_Zhou_2023}. Their capacity to capture global context makes them suitable for complex volumetric inputs such as DCE-MRI. However, their implementation in breast cancer imaging is hindered by the absence of large, standardized datasets.

To address these challenges, we present BreastDCEDL, a comprehensive deep learning-ready dataset for breast DCE-MRI. By integrating data from I-SPY2~\cite{Wang_Yee_2019} (985 patients), Duke~\cite{Ashirbani_2021} (922 patients) and I-SPY-1~\cite{Newitt_Hylton_2016} (221 patients), we have created a standardized resource of unprecedented scale. The data set transforms complex digital imaging and communications in medicine (DICOM) structures into accessible 3D arrays with preserved signal integrity, unified tumor annotations, and harmonized clinical metadata.

Using this resource, we implemented a Vision Transformer (ViT) workflow for DCE-MRI analysis. Our approach demonstrates the feasibility of applying transformer architectures to this complex imaging modality, achieving state-of-the-art pCR prediction. This implementation is the first successful application of transformer models to breast DCE-MRI data.

Through BreastDCEDL and our transformer-based modeling workflow, our objective is to accelerate progress in AI-assisted breast cancer care by providing the research community with the resources needed to develop, validate, and compare deep learning models in a standardized environment.

\section{Methods}

\subsection{BreastDCEDL Dataset Composition}

BreastDCEDL integrates three datasets from The Cancer Imaging Archive (TCIA)\cite{Clark_2013}: I-SPY2 (Investigation of Serial Studies to Predict Your Therapeutic Response with Imaging And moLecular Analysis 2)\cite{Li_Newitt_Gibbs_Wilmes_Jones_Arasu_Strand_Onishi_Nguyen_Kornak_et_al._2020, Wang_Yee_2019}, I-SPY1 (QuantumLeap Healthcare Collaborative)\cite{Hylton_Gatsonis_Rosen_Lehman_Newitt_Partridge_Bernreuter_Pisano_Morris_Weatherall_et_al._2015, Newitt_Hylton_Archive}, which is similar to I-SPY2, and Duke\cite{Ashirbani_2021}. I-SPY2 contains 985 patients, 3 of whom had missing clinical data. In I-SPY1, several cases have missing DCE-MRI data, with only 172 patients having at least three scans. Duke contains 922 patients, 6 of whom had missing Dynamic Contrast-Enhanced (DCE) data. This integration resulted in a cohort of 2,070 patients (Table~\ref{tab:demographics}). This integration resulted in a cohort of 2,070 patients (Table~\ref{tab:demographics}).

\begin{table}[ht]
\centering
\begin{threeparttable}
\caption{Patient demographic characteristics and biomarker statuses in the BreastDCEDL database.}
\label{tab:demographics}
\vspace{0.3cm}
\footnotesize
\begin{tabular}{@{}llcccc@{}}
\toprule
\textbf{Section} & \textbf{\makecell{Characteristic}} & \textbf{\makecell{Duke\\(n=916)}} & \textbf{\makecell{I-SPY1\\(n=172)}} & \textbf{\makecell{I-SPY2\\(n=982)}} & \textbf{\makecell{Overall\\(n=2070)}} \\
\midrule
\multicolumn{6}{l}{\textbf{Demographics}} \\
& Age$^{\dagger}$ (years)      & 52.4           & 48.0           & 48.8           & 50.3 \\
& Race: Black (\%)             & 200 (21.8)     & 28 (16.3)      & 117 (11.9)     & 345 (16.7) \\
& Race: White (\%)             & 648 (70.7)     & 123 (71.5)     & 785 (79.9)     & 1556 (75.2) \\
& Total patients (\%)          & 916 (100.0)    & 172 (100.0)    & 982 (100.0)    & 2070 (100.0) \\

\addlinespace
\multicolumn{6}{l}{\textbf{HR Status}} \\
& HR+ (\%)                     & 696 (76.0)     & 95 (55.9)      & 536 (54.6)     & 1327 (64.2) \\
& HR$^-$ (\%)                  & 220 (24.0)     & 75 (44.1)      & 446 (45.4)     & 741 (35.8) \\

\addlinespace
\multicolumn{6}{l}{\textbf{Receptor Subtype}} \\
& HER2+ (\%)                   & 163 (17.8)     & 53 (30.8)      & 242 (24.6)     & 458 (22.1) \\
& HR+/HER2$^-$ (\%)            & 592 (64.6)     & 66 (38.4)      & 381 (38.8)     & 1039 (50.2) \\
& Triple Negative (\%)         & 161 (17.6)     & 43 (25.0)      & 359 (36.6)     & 563 (27.2) \\

\addlinespace
\multicolumn{6}{l}{\textbf{pCR}} \\
& pCR+ (\%)                    & 63 (21.1)      & 49 (28.5)      & 316 (32.2)     & 428 (29.5) \\
& pCR$^-$ (\%)                 & 235 (78.9)     & 123 (71.5)     & 666 (67.8)     & 1024 (70.5) \\
\bottomrule
\end{tabular}
\vspace{0.3cm}
\begin{tablenotes}
\small
\item \textbf{Notes:} HR = Hormone Receptor (ER and/or PR positive); HER2+ = HER2 positive regardless of HR status; HR+/HER2$^-$ = Hormone receptor positive and HER2 negative; Triple Negative = ER$^-$/PR$^-$/HER2$^-$; pCR = pathological complete response.
\item $^{\dagger}$ Age is reported as mean.
\item Values in parentheses indicate percentages.
\item Percentages for pCR are based on patients with available pCR data (Duke: 298, I-SPY1: 172, I-SPY2: 982, Overall: 1452).
\end{tablenotes}
\end{threeparttable}
\end{table}

\subsection{DCE-MRI Data Standardization}

The raw data from I-SPY2, I-SPY1, and Duke were collected from multiple centers; I-SPY2 alone contributed data from over 22 centers. We extracted slice location and acquisition time from DICOM metadata and organized the data temporally by acquisition time and spatially along anatomical axes. The datasets stored information, such as modality type, contrast injection time, and scan intervals, across different DICOM tags with diverse filename formats and metadata descriptions, requiring systematic reconciliation.

The original DICOM data used 16-bit integer or 16-bit unsigned integer representation; however, we preserved the full dynamic range by converting to 64-bit floating-point format during standardization. We organized the 3D files with a clear naming convention following the pattern \texttt{<dataset>\_<patient id>\_acq<acquisition number>}. To facilitate validation and exploration of the MRI data, we provide a mapping table linking the raw DICOM slices to the 3D NIfTI planes in the project's Git repository, and a comprehensive DICOM tag dictionary.

\subsection{Tumor Annotation Processing and Integration}

The I-SPY trial databases contain processed data in addition to raw MRI slices, including tumor segmentation, Region of Interest (ROI) delineation, Selected Enhancement Ratio (SER) images (pre-contrast, early post-contrast, and late post-contrast), enhancement calculation maps, and analysis masks that encode the segmentation process. We converted these analysis masks into binary 3D files, coding tumor regions as 1 and background tissue as 0. The masks were oriented along the same axis as the original MRI scans to facilitate precise alignment.

For the Duke dataset, annotations were created based on the pre-contrast Dynamic Contrast-Enhanced Magnetic Resonance (DCE-MR) images, the first post-contrast images, and the difference between the two. Expert radiologists annotated bounding boxes on planes containing the largest tumor area. We transformed these bounding box coordinates to correspond with our standardized DCE-MRI alignment. MRI scans across all datasets can contain multiple tumors (up to 9 per scan in the Duke dataset); however, only the largest (primary) tumor was annotated.

\subsection{Clinicopathologic Data Harmonization}

We standardized heterogeneous clinical metadata across the three source datasets, unifying common demographic features—age at diagnosis, race, and menopausal status—and tumor characteristics—Hormone Receptor (HR) and Human Epidermal Growth Factor Receptor 2 (HER2) status. HR status determination varied by institutional thresholds: I-SPY2 used $\geq$1\% staining (most sensitive), Duke employed Allred score $>$3 (intermediate), and I-SPY1 applied $\geq$10\% staining (most stringent). Consequently, I-SPY2 classified the greatest proportion of tumors as HR+, while I-SPY1 classified the fewest. For HER2 status, Duke used a FISH ratio $>$2.2, while both I-SPY trials used $\geq$2.0, resulting in marginally fewer tumors classified as HER2+ in the Duke dataset. 

We calculated tumor volumes for the I-SPY datasets by multiplying malignant voxel counts by physical voxel dimensions, whereas the Duke dataset provided tumor volumes pre-categorized into five size classes (1--5 cm$^3$). We excluded patients with fewer than three MRI acquisitions and organized all clinicopathologic and imaging data in a standardized CSV format, documenting scan parameters, physical dimensions, SER timepoints, and clinical variables. Population characteristics are summarized in Table~\ref{tab:demographics}.

\subsection{Vision Transformer for pCR Prediction}

\begin{figure}[!htbp]
\centering
\includegraphics[width=0.95\textwidth]{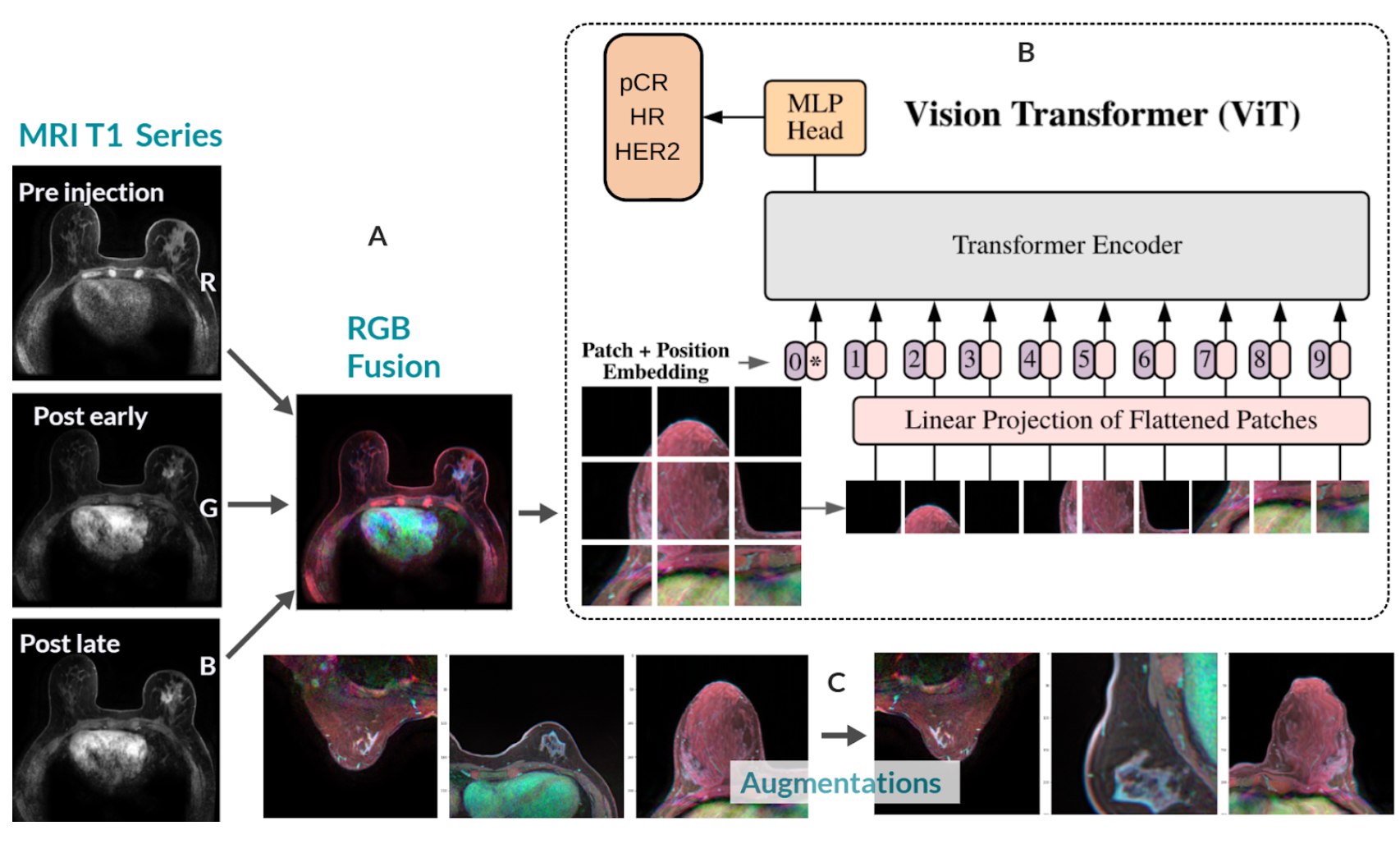}
\caption{DCE-MRI fusion input for Vision Transformer (ViT) classification. (A) Images from pre-contrast, early post-contrast, and late post-contrast phases are assigned to red, green, and blue channels and fused into a 2D RGB image. (B) The ViT model processes RGB images by dividing them into patches, embedding these as tokens, and passing them through a BERT-like architecture. (C) Examples of image augmentation techniques used during model training. Panel B adapted from~\cite{Dosovitskiy_et_al_2020}.}
\label{fig:vit_pipeline}
\end{figure}

We implemented a Vision Transformer (ViT) approach for predicting pathologic complete response (pCR) from DCE-MRI data. Input preparation involved generating RGB-fused images from DCE-MRI slices containing tumor regions, with pre-contrast, early post-contrast, and late post-contrast acquisitions assigned to the red, green, and blue channels, respectively, as illustrated in Figure~\ref{fig:vit_pipeline}. 

For the I-SPY trials, we utilized Signal Enhancement Ratio (SER) timepoints 0, 2, and the minimum between the last index and 6. For the Duke dataset, timepoints 0, 1, and the final acquisition were used. All images underwent Min-Max normalization and conversion to 8-bit format for compatibility with pre-trained models.

\subsection{Benchmark Establishment}

We created fixed Train–Validation–Test partitions for the BreastDCEDL dataset to ensure reproducible benchmarking across studies. We randomly split patients with the pathologic complete response (pCR) status available according to the response rate and distributed the remaining patients based on the status of HER2. We redistributed a few randomly selected patients between groups based on pathologic complete response, HR, and HER2 status to maintain balanced biomarker representation across all dataset partitions as much as possible.

The final dataset partitions comprised 1,099 patients (29.3\% pCR rate) for training, 177 patients (29.9\% pCR rate) for validation, and 176 patients (30.1\% pCR rate) for testing, with an overall pCR rate of 29.5\% across all 1,452 patients with available pCR data. These standardized splits are provided with the data set to facilitate the direct comparison of different modeling approaches and ensure reproducible research results.

\section{Results}

\subsection{BreastDCEDL Dataset}

\begin{figure}[!htbp]
\centering
\includegraphics[width=0.95\textwidth]{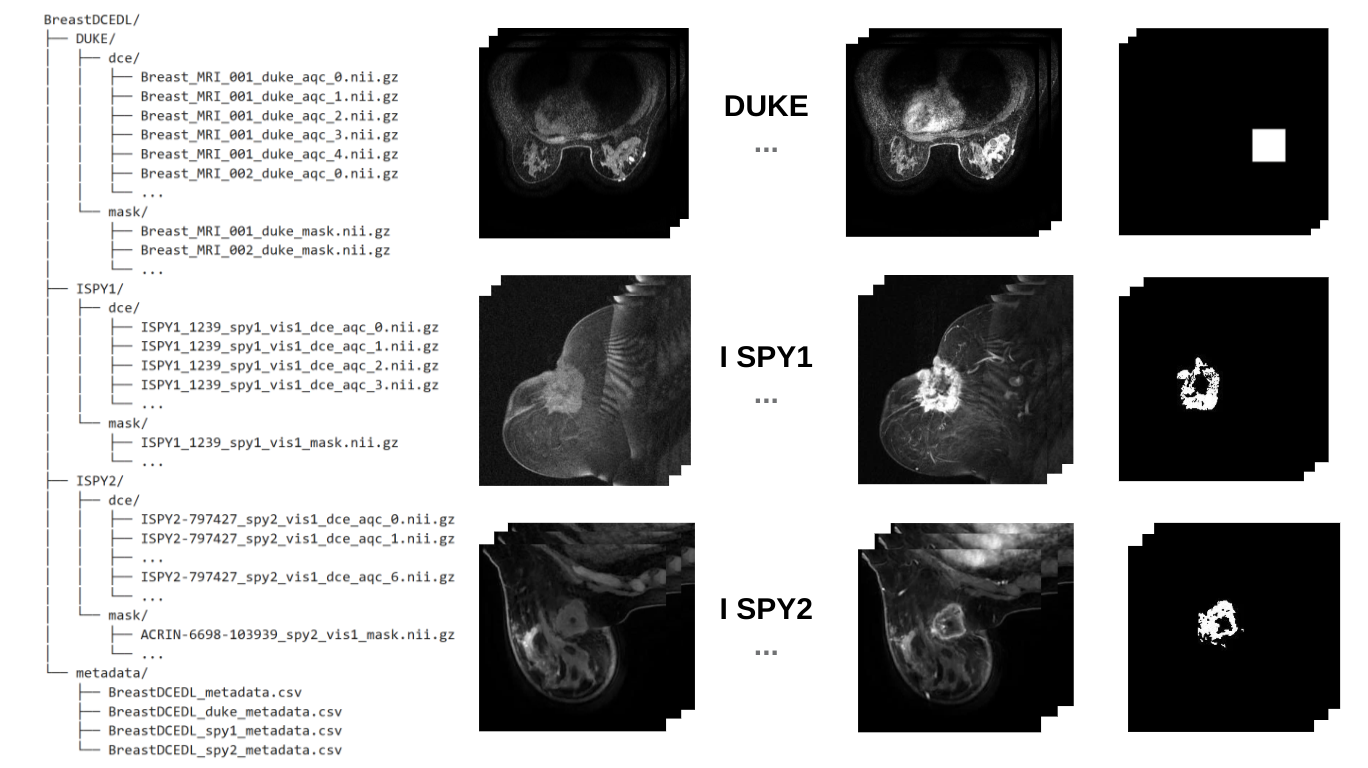}
\caption{BreastDCEDL structure and imaging examples. Standardized 3D NIfTI format with clinical annotations facilitates organized data integration. Representative DCE-MRI examples from contributing cohorts illustrate diverse imaging characteristics across institutions.}
\label{fig:dataset_structure}
\end{figure}

We developed BreastDCEDL, a comprehensive and standardized DCE-MRI dataset designed for deep learning applications and computational analysis, including radiomics. The data set consolidates data from three major clinical cohorts: Duke (n = 916), I-SPY1 (n = 172) and I-SPY2 (n = 982), creating a resource of 2,070 patients with preserved original DICOM intensity values, spatial resolution, and imaging parameters. All data have been converted to the standardized 3D NIfTI format with complete clinical annotations and ready-to-use train, validation, and test partitions. The data organization structure and representative examples from each cohort are presented in Figure~\ref{fig:dataset_structure}.

\subsection{Benchmark Partitions}

Our benchmark partitions maintain consistent pathologic complete response (pCR) rates across the Train-Validation-Test splits: 29. 3\%, 29. 9\%, and 30. 1\%, respectively, among the 1,452 patients with available response data. This design enables fair and reproducible comparisons between computational methods while preserving the clinical heterogeneity essential for robust model development.

\subsection{Vision Transformer Performance}

Our Vision Transformer (ViT) implementation demonstrated strong performance in predicting pathologic complete response (pCR), achieving results that compare favorably with previous studies (Table~\ref{tab:performance_results}). The model performed particularly well on the I-SPY2 dataset, achieving an area under the receiver operating characteristic curve (AUC) of 0.85 overall and 0.94 for HR+/HER2$^-$ tumors, with 93\% accuracy and perfect precision. 

This performance exceeds that of Li et al.~\cite{Li_Newitt_Gibbs_Wilmes_Jones_Arasu_Strand_Onishi_Nguyen_Kornak_et_al._2020}, who reported an AUC of 0.83 using traditional machine learning on DCE-MRI features from 384 I-SPY2 patients. It also aligns with Dammu et al.~\cite{Dammu_Ren_Duong_2023}, who achieved an AUC of 0.70 using RGB fusion with a residual CNN on 155 I-SPY1 patients, while our model achieved an AUC of 0.68 on 172 unfiltered I-SPY1 patients. 

Our results further improve upon the recent large-scale study by Li et al.~\cite{Li_Onishi_Gibbs_Wilmes_Le_Metanat_Price_Joe_Kornak_Yau_et_al._2025}, who analyzed 814 I-SPY2 patients and reported AUC values ranging from 0.68 to 0.74 across receptor subtypes, with HR+/HER2$^-$ showing the highest performance—consistent with our findings.

\begin{table}[ht]
\centering
\begin{threeparttable}
\caption{Pathologic complete response (pCR) prediction results using radiomic features and machine learning.}
\label{tab:performance_results}
\vspace{0.3cm}
\footnotesize
\begin{tabular}{@{}llcccccccc@{}}
\toprule
\textbf{Group} & \textbf{Subtype} & \textbf{N} & \textbf{\makecell[c]{pCR\\Rate}} & \textbf{Acc.} & \textbf{AUC} & \textbf{Sens.} & \textbf{Spec.} & \textbf{Prec.} & \textbf{NPV} \\
\midrule
\multicolumn{10}{l}{\textbf{Overall Test Set}} \\
& Overall & 177 & 30.1\% & 0.75 & 0.72 & 0.27 & 0.95 & 0.70 & 0.75 \\
& HR$^-$/HER2$^-$ & 63 & 36.5\% & 0.65 & 0.63 & 0.26 & 0.88 & 0.55 & 0.67 \\
& HER2+ & 43 & 42.0\% & 0.72 & 0.68 & 0.44 & 0.92 & 0.80 & 0.70 \\
& HR+/HER2$^-$ & 69 & 16.0\% & 0.86 & 0.64 & 0.10 & 1.00 & 1.00 & 0.86 \\

\addlinespace
\multicolumn{10}{l}{\textbf{By Source Dataset}} \\
& I-SPY1 & 35 & 34.3\% & 0.72 & 0.68 & 0.17 & 1.00 & 1.00 & 0.70 \\
& I-SPY2 & 99 & 32.3\% & 0.73 & 0.78 & 0.28 & 0.94 & 0.70 & 0.73 \\
& Duke & 43 & 21.0\% & 0.77 & 0.54 & 0.22 & 0.91 & 0.40 & 0.82 \\

\addlinespace
\multicolumn{10}{l}{\textbf{I-SPY2 with Clinical Features}} \\
& Overall & 99 & 32.3\% & 0.76 & 0.85 & 0.34 & 0.96 & 0.79 & 0.75 \\
& HR$^-$/HER2$^-$ & 40 & 47.5\% & 0.68 & 0.75 & 0.32 & 1.00 & 1.00 & 0.62 \\
& HER2+ & 19 & 42.1\% & 0.78 & 0.84 & 0.75 & 0.82 & 0.75 & 0.82 \\
& HR+/HER2$^-$ & 40 & 12.5\% & 0.93 & 0.94 & 0.41 & 1.00 & 1.00 & 0.92 \\
\bottomrule
\end{tabular}
\vspace{0.2cm}
\begin{tablenotes}
\small
\item \textbf{Abbreviations:} pCR = pathologic complete response; NPV = negative predictive value; HR = Hormone Receptor; HER2+ = HER2 positive; Acc. = Accuracy; Sens. = Sensitivity; Spec. = Specificity; Prec. = Precision
\item \textbf{Notes:} HR+/HER2$^-$ = Hormone receptor positive and HER2 negative; Triple Negative = ER$^-$/PR$^-$/HER2$^-$. I-SPY2 results include clinical, demographic, and tumor features.
\end{tablenotes}
\end{threeparttable}
\end{table}

For HER2+ tumors, our model achieved an AUC of 0.68 overall on BreastDCEDL and 0.84 on the I-SPY2 cohort, which is comparable to Mohamed et al.~\cite{Mohamed_Panthi_Adrada_Boge_Candelaria_Chen_Guirguis_Hunt_Huo_Hwang_et_al._2024}, who reported AUCs of 0.77 and 0.85 on small validation cohorts (n=28 and n=20) using pre-trained CNN with SVM approaches. Performance for triple-negative tumors remained challenging but competitive, with our overall AUC of 0.63 and I-SPY2 AUC of 0.75, aligning with current literature standards where Li et al.~\cite{Li_Onishi_Gibbs_Wilmes_Le_Metanat_Price_Joe_Kornak_Yau_et_al._2025} achieved less than 0.63 AUC on a single-center cohort of 163 triple-negative cases.

Several studies have incorporated MRI sequences beyond DCE-MRI to enhance predictive performance. Liu et al.\cite{Liu_Li_Qu_Zhang_Zhou_Li_Sun_Tang_Jiang_Li_et_al._2019} achieved an AUC of 0.86 for HR+/HER2$^-$ tumors using machine learning on features extracted from DCE-MRI, T2-weighted imaging, and diffusion-weighted imaging across 586 patients from four hospitals. Similarly, Janssen et al.\cite{Janssen_Janse_Vries_Velden_Ben_Bosch_Sartori_Jovelet_Agterof_Huinink_et_al._2024} utilized T2-weighted and T1-weighted sequences alongside DCE-MRI, achieving an AUC of 0.86 on a private single-center dataset of 61 patients. While these multimodal approaches demonstrate the value of incorporating diverse imaging information, our DCE-MRI-only method achieved comparable performance using a single imaging modality, suggesting that transformer architectures effectively capture the complex spatiotemporal patterns inherent in dynamic contrast-enhanced sequences. This indicates that substantial improvements may be achievable when additional modalities are incorporated into BreastDCEDL.

\begin{figure}[!htbp]
\centering
\includegraphics[width=0.8\textwidth]{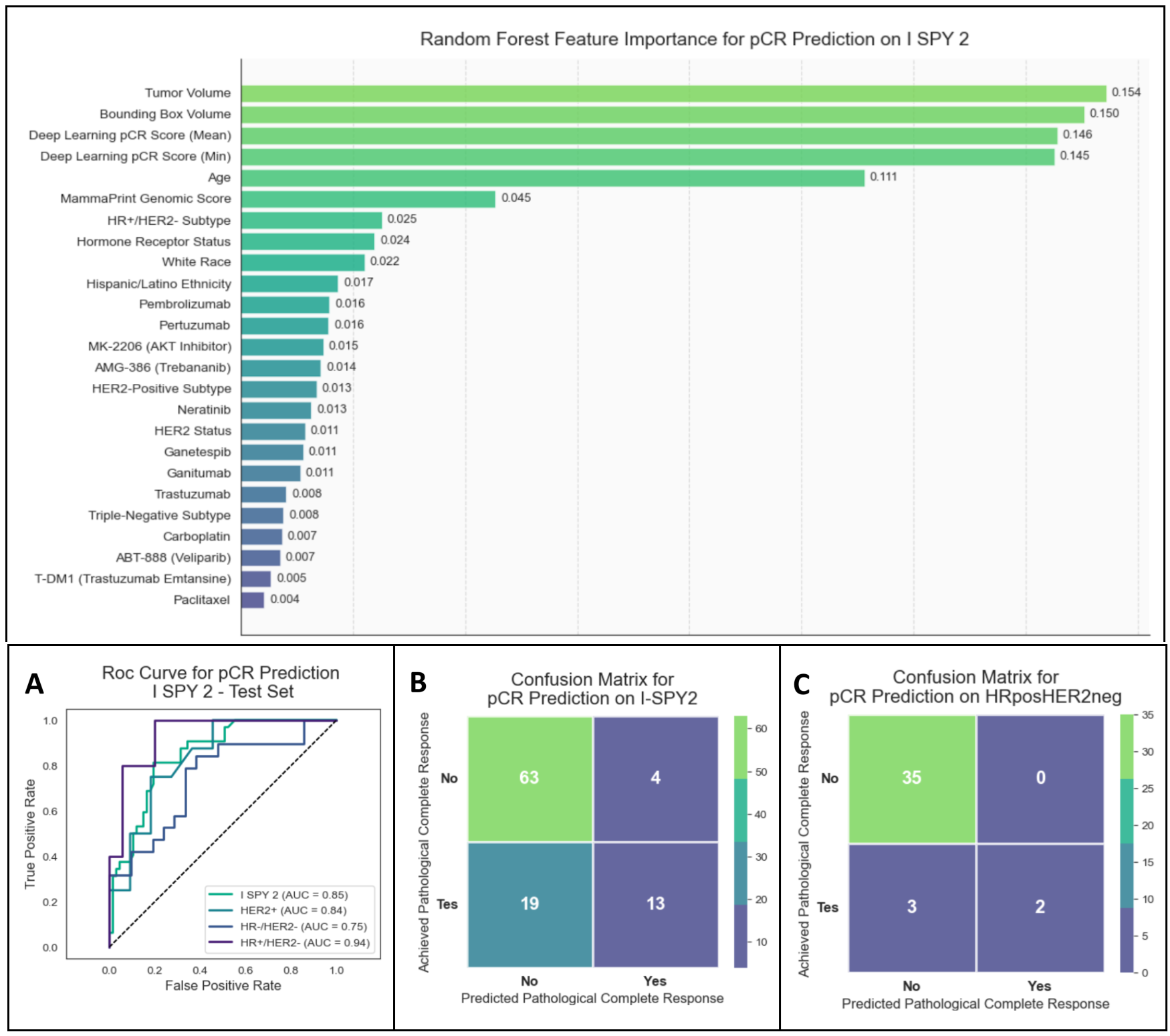}
\caption{Random Forest model performance for pathological complete response (pCR) prediction in the I-SPY2 cohort. Top panel shows feature importance rankings for pCR prediction. Bottom panels show: \textbf{A} Receiver operating characteristic (ROC) curves for overall cohort and hormone receptor subtypes, \textbf{B} Confusion matrix showing model classification performance on I-SPY2, and \textbf{C} Confusion matrix showing model classification performance on HR+/HER2$^-$ subtype of I-SPY2 cohort.}
\label{fig:feature_importance}
\end{figure}

Feature importance analysis revealed that tumor volume consistently ranked as the most critical predictor across all tested models, with tumor volume, bounding box volume, and deep learning-derived features appearing at the top of importance rankings (Figure~\ref{fig:feature_importance}). Age typically ranked high in predictive importance, followed by hormone receptor status, while race consistently ranked lowest among clinical variables. These findings underscore the complementary value of quantitative imaging biomarkers and clinical parameters in treatment response prediction.

\section{Discussion}

We presented BreastDCEDL, a comprehensive, standardized, and publicly available DCE-MRI dataset for breast cancer, comprising 2,070 patients across three major cohorts. This resource addresses critical barriers that have impeded applying deep learning methods to breast cancer imaging. BreastDCEDL preserves original DICOM acquisition parameters without imposed preprocessing, providing comprehensive metadata, including spatial resolutions, slice thicknesses, and scanner specifications. This approach enables researchers to develop and evaluate their own normalization and harmonization methods, a fundamental requirement for the advancement of the MRI research methodology.

Our Vision Transformer model achieved competitive performance with published benchmarks, demonstrating robust discriminative capacity. Across 177 test cases, the model achived 75\% accuracy and 0.72 AUC (Table~\ref{tab:performance_results}). The model exhibited high specificity (0.95) with corresponding low sensitivity (0.27), reflecting a conservative prediction profile that minimizes false positives. This performance characteristic aligns with clinical priorities, where accurate identification of nonresponders provides critical pretreatment guidance, while false positives carry lower clinical risk given that patients proceed to neoadjuvant chemotherapy regardless. For HR+/HER2$^-$ tumors in the I-SPY2 cohort with integrated clinical data (N=40, 12.5\% pCR rate), our approach achieved exceptional performance (AUC 0.94, accuracy 93\%, NPV 92\%), representing state-of-the-art results that could inform treatment decisions.

Several limitations warrant mention. We observed performance variations across data sets (I-SPY2: AUC 0.78; Duke: AUC 0.54; I-SPY1: AUC 0.68), likely reflecting the larger sample size of I-SPY2, superior image quality, and tumor volumes. Additional factors include the diversity of the imaging protocol and our preservation of raw, unprocessed data; while methodologically valuable for harmonization research, this approach may reduce performance consistency compared to normalized datasets. 

Our methodological choices impose constraints. The 8-bit representation and RGB fusion approach, while enabling pre-trained model leverage, may obscure DCE-MRI's spatiotemporal patterns. The RGB fusion method blurs specific enhancement patterns from individual timepoints, which may be critical for biomarker prediction. Furthermore, our 2D implementation fails to capture the 3D spatio-temporal information in DCE-MRI sequences, representing a limitation for comprehensive tumor characterization.

The Vision Transformer architecture presents additional constraints. Fixed, nonoverlapping patches, while suitable for natural images, limit analysis of fine-grained medical imaging details. Future implementations could benefit from architectures like Shifted Window Transformers~\cite{Liu_Lin_Cao_Hu_Wei_Zhang_Lin_Guo_2021}, that better handle medical imaging's spatial granularity requirements. Furthermore, our two-stage approach, combining deep learning features with clinical variables through traditional machine learning, could be superseded by end-to-end transformer architectures directly incorporating clinical data within the deep learning framework.

Transformer models offer more than powerful representation learning; they generate attention maps that may provide insights into DCE-MRI interpretation. These attention patterns could identify novel imaging biomarkers in breast MRI, improving our understanding of treatment response prediction. By making BreastDCEDL publicly available alongside code and pre-trained models, we aim to accelerate progress in AI-assisted breast cancer care and enable more advanced architectural innovations to address current limitations.

\section{Data and Code Availability}

The code for data processing and model implementation is available on GitHub: \url{https://github.com/naomifridman/BreastDCEDL}. Processed data sets including the I-SPY1 subset, Duke tumor-centered cohort samples, and pre-trained model weights are deposited in Zenodo~\cite{Fridman_Solway_Fridman_Barnea_2025} (\url{https://doi.org/10.5281/zenodo.15627233}). Complete imaging data sets from the I-SPY1, I-SPY2 and Duke cohorts can be accessed through The Cancer Imaging Archive (TCIA) and converted to standardized NIfTI format using our provided processing pipeline. Standardized train-validation-test splits and comprehensive metadata are included to ensure reproducible model development and evaluation.

\end{document}